\documentclass[twocolumn]{article}

\usepackage[twoside,
  paperwidth=210mm,
  paperheight=280mm,
  textheight=693pt,
  textwidth=522pt,
  inner=15mm,
  top=15mm,
  headheight=50pt,
  headsep=10pt,
  footskip=12pt,
  footnotesep=28pt plus 2pt minus 6pt,
  columnsep=18pt]{geometry}
\usepackage{url}
\usepackage{algorithm}
\usepackage{algorithmic}
\usepackage{eqparbox}
\usepackage{amsmath,amssymb}
\usepackage{multicol}
\usepackage{subfigure}
\usepackage{graphicx}
\usepackage{color}
\usepackage{xcolor}
\usepackage{array}
\usepackage{multirow}
\usepackage[nolist,nohyperlinks]{acronym}
\acrodef{GDPR}{\emph{General Data Protection Regulation}}
\acrodef{EU}{\emph{European Union}}
\acrodef{EEA}{\emph{European Economic Area}}
\acrodef{FedAvg}{Federated Averaging}
\acrodef{FedBoosting}{Federated Boosting}
\acrodef{FL}{Federated Learning}
\acrodef{GAN}{Generative Adversarial Network}
\acrodef{DL}{Deep Learning}
\acrodef{HE}{Homomorphic Encryption}
\acrodef{DP}{Differential Privacy}
\acrodef{CRNN}{Convolutional Recurrent Neural Network}
\acrodef{CNN}{Convolutional Neural Network}
\acrodef{GPU}{Graphics Processing Unit}
\acrodef{Non-IID}{Non-Independent and Non-Identically Distributed}
\acrodef{IID}{Independent and Identically Distributed}
\acrodef{EMD}{Earth Mover’s Distance}
\acrodef{MPC}{Secure Multi-Party Computation}
\acrodef{BiLSTM}{Bidirectional Long-Short Term Memory}
\acrodef{CTC}{Connectionist Temporal Classification}
\acrodef{ZKP}{Zero-Knowledge Proof}
\acrodef{GRNN}{Generative Regression Neural Network}

\newcommand\blfootnote[1]{%
  \begingroup
  \renewcommand\thefootnote{}\footnote{#1}%
  \addtocounter{footnote}{-1}%
  \endgroup
}

\begin{document}

\title{FedBoosting: Federated Learning with Gradient Protected Boosting for Text Recognition\blfootnote{First Author: H. Ren (hanchi.ren@swansea.ac.uk). Corresponding Authors: J. Deng (jingjing.deng@durham.ac.uk) and X. Xie (x.xie@swansea.ac.uk). Co-Authors: X. Ma (xkma@xidian.edu.cn) and Y. Wang (chuan@xaut.edu.cn).}}
% \footnote{The paper is under consideration at Pattern Recognition Letters.}}

\author{Hanchi Ren, Jingjing Deng*, Xianghua Xie*, Xiaoke Ma, Yichuan Wang}

\date{}
\maketitle

\begin{abstract}
Conventional machine learning methodologies require the centralization of data for model training, which may be infeasible in situations where data sharing limitations are imposed due to concerns such as privacy and gradient protection. The Federated Learning (FL) framework enables the collaborative learning of a shared model without necessitating the centralization or sharing of data among the data proprietors. Nonetheless, in this paper, we demonstrate that the generalization capability of the joint model is suboptimal for Non-Independent and Non-Identically Distributed (Non-IID) data, particularly when employing the Federated Averaging (FedAvg) strategy as a result of the weight divergence phenomenon. Consequently, we present a novel boosting algorithm for FL to address both the generalization and gradient leakage challenges, as well as to facilitate accelerated convergence in gradient-based optimization. Furthermore, we introduce a secure gradient sharing protocol that incorporates Homomorphic Encryption (HE) and Differential Privacy (DP) to safeguard against gradient leakage attacks. Our empirical evaluation demonstrates that the proposed Federated Boosting (FedBoosting) technique yields significant enhancements in both prediction accuracy and computational efficiency in the visual text recognition task on publicly available benchmarks.
\end{abstract}

%% main text
\section{Introduction}
\label{sec:introduction}

The protection of personal data and preservation of privacy have garnered considerable interest from researchers in recent years~\cite{aono2017privacy,hesamifard2018privacy,ryffel2018generic,al2019privacy,liu2020survey,tanuwidjaja2020privacy,koti2021swift}. Traditional machine learning methodologies necessitating centralized data for model training may be infeasible due to data sharing restrictions. Consequently, decentralized data-training approaches are increasingly appealing, as they offer advantages in preserving privacy and ensuring data security. \ac{FL}~\cite{konevcny2016federated,mcmahan2017communication} was introduced to address these concerns by enabling individual data providers to collaboratively train a shared global model without the need for central data aggregation. McMahan \emph{et al.} \cite{mcmahan2017communication} proposed a practical decentralized training method for deep networks based on averaging aggregation. Empirical studies conducted on various datasets and architectures demonstrated the robustness of \ac{FL} for handling unbalanced and \ac{IID} data. While frequent updating generally leads to improved prediction performance, communication costs can increase substantially, particularly for large datasets~\cite{kairouz2019advances,konevcny2016federatedlearning,mcmahan2017communication,li2018federated,smith2017federated}. To address the efficiency issue, Kone{\v{c}}n{\`y} \emph{et al.} ~\cite{konevcny2016federatedlearning} proposed two weight updating methods based on \ac{FedAvg}, namely structured updates and sketched updates approaches, to reduce the up-link communication costs associated with transmitting gradients from local machines to a centralized server.

\begin{figure*}[ht!]
\centering
 \begin{center}
  \includegraphics[width=0.70\linewidth]{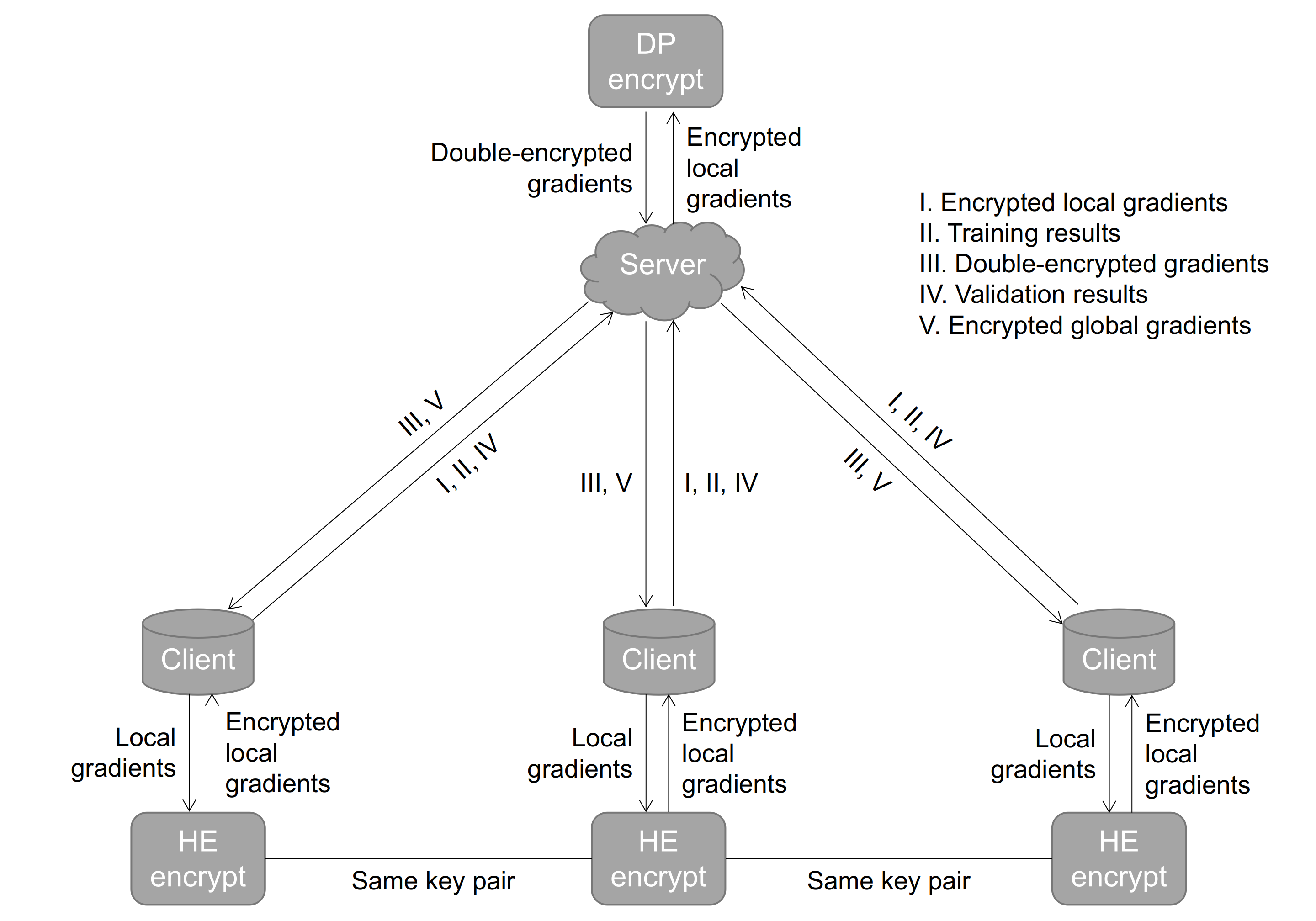}
 \end{center}
\caption{The schematic diagram illustrates the proposed \ac{FedBoosting} and encryption protocol. For demonstration purposes, two clients are depicted; however, the proposed method is designed to accommodate an arbitrary number of local clients.}
\label{fig:FedBoost_illustration}
\end{figure*}

Prediction performance and data privacy represent two primary challenges in \ac{FL} research. On the one hand, the accuracy of \ac{FL} is known to decline considerably when applied to \ac{Non-IID} data~\cite{zhao2018federated,arivazhagan2019federated,liang2020think,lin2020ensemble,li2021decentralized,li2019fedmd,peng2019federated,ghosh2020efficient,briggs2020federated}. Zhao \emph{et al.}~\cite{zhao2018federated} demonstrated that weight divergence can be quantitatively measured using \ac{EMD} between class distributions on each local machine and the global population distribution. Consequently, they proposed sharing a small subset of data among all edge devices to enhance model generalization on \ac{Non-IID} data. However, this strategy may not be viable when data sharing restrictions are in place, potentially leading to privacy breaches. Li \emph{et al.}~\cite{li2019convergence} examined the convergence properties of \ac{FedAvg} and identified a trade-off between its communication efficiency and convergence rate, arguing that the model converges slowly on heterogeneous datasets. Our empirical study in this paper confirms that, given \ac{Non-IID} datasets, the training process requires considerably more iterations to reach an optimal solution and often fails to converge, particularly when local models are trained on large-scale datasets with a small batch size, or the global model is aggregated after a substantial number of epochs. The feasibility of adaptive aggregation by leveraging training and testing outcomes has been established~\cite{chen2020dealing,pejo2020quality}. On the other hand, model gradients are generally considered safe to share in \ac{FL} systems for model aggregation. Nonetheless, some studies have revealed the feasibility of recovering training data information from model gradients. For instance, Fredrikson \emph{et al.} \cite{fredrikson2015model} and Melis \emph{et al.} \cite{melis2019exploiting} reported two methods capable of identifying a sample with specific properties in the training batch. Hitaj \emph{et al.}~\cite{hitaj2017deep} proposed a \ac{GAN} model as an adversarial client for estimating data distribution from the outputs of other clients without access to their training data. Zhu \emph{et al.}~\cite{zhu2019deep} and Zhao \emph{et al.}~\cite{zhao2020idlg} demonstrated that data recovery can be framed as a gradient regression problem, assuming the gradient from a targeted client is available, which is a largely valid assumption in most \ac{FL} systems. Moreover, the \ac{GRNN} proposed by Ren \emph{et al.}~\cite{ren2022grnn} comprises two branches of generative models: one based on \ac{GAN} for generating fake training data and the other on a fully-connected layer for generating corresponding labels. The training data is revealed by regressing the true gradient and the fake gradient generated by the fake data and relevant label.

In this paper, we propose \ac{FedBoosting} method to address the weight divergence and gradient leakage issues in general \ac{FL} framework. Instead of treating individual local models equally when the global model is aggregated, we consider the data diversity of local clients in terms of the status of convergence and the ability of generalization. To address the potential risk of data leakage via shared gradients, a \ac{DP} based linear aggregation method is proposed using \ac{HE}~\cite{paillier1999public} to encrypt the gradients which provides two layers of protection. The proposed encryption scheme only leads to a negligible increase in computational cost.

In this paper, we introduce the \ac{FedBoosting} method to address weight divergence and gradient leakage issues within the general \ac{FL} framework. Rather than treating individual local models equally during global model aggregation, we take into account the data diversity of local clients concerning their convergence status and generalization ability. To mitigate the potential risk of data leakage via shared gradients, we propose a \ac{DP}-based linear aggregation method using \ac{HE}~\cite{paillier1999public} to encrypt gradients, providing dual layers of protection. The proposed encryption scheme incurs only a negligible increase in computational cost.

We evaluate the proposed method on text recognition task using public benchmarks, as well as a binary classification task on two datasets, demonstrating its superiority in terms of convergence speed, prediction accuracy, and security. We also assess the performance reduction due to encryption. Our contributions can be summarized in four main points:

\begin{itemize}
\item We propose a novel aggregation strategy, \ac{FedBoosting}, for \ac{FL} to address weight divergence and gradient leakage issues. Empirically, we demonstrate that \ac{FedBoosting} converges considerably faster than \ac{FedAvg}, while maintaining communication costs comparable to traditional approaches. Particularly when local models are trained with small batch sizes and the global model is aggregated after a large number of epochs, our approach can still converge to a reasonable optimum, whereas \ac{FedAvg} often fails in such cases.
\item We introduce a dual-layer protection scheme utilizing \ac{HE} and \ac{DP} to encrypt gradients exchanged between servers and clients, safeguarding data privacy against gradient leakage attacks.
\item We demonstrate the feasibility of our method on two datasets by visually evaluating decision boundaries. Additionally, we showcase its superior performance in a visual text recognition task on multiple large-scale \ac{Non-IID} datasets compared to centralized approaches and \ac{FedAvg}. The experimental results confirm that our approach outperforms \ac{FedAvg} in terms of convergence speed and prediction accuracy, suggesting that the \ac{FedBoosting} strategy can be integrated with other \ac{DL} models in privacy-preserving scenarios.
\item We provide a publicly available implementation of the proposed \ac{FedBoosting} method to ensure reproducibility. The implementation can also be executed in a distributed, multi-\acp{GPU} setup.\footnote{\url{https://github.com/Rand2AI/FedBoosting}}
\end{itemize}

The remainder of this paper is organized as follows: In Section~\ref{sec:rw}, we present related work on encryption methods, collaborative learning, and gradient leakage. The proposed \ac{FedBoosting} method and corresponding encryption techniques are described in Section~\ref{sec:pm} and evaluated using a text recognition task and a binary classification task. Section~\ref{sec:er} provides details of the experiments, discussions on the results, and a performance comparison. Finally, we draw conclusions in Section~\ref{sec:cc}.

%=====================================================================

\section{Related Work}
\label{sec:rw}

\ac{FL} has been proposed for privacy-preserving machine learning to train models across multiple decentralized edge devices or clients containing local data samples \cite{mcmahan2017communication,konevcny2016federatedlearning,konevcny2016federated,mcmahan2017federated,yang2019federated,zhang2020fedocr}. The \ac{FL} framework retains raw data with the owners and trains models locally at individual client nodes, while gradients from these models are exchanged and aggregated rather than the data. Compared to \ac{MPC}~\cite{yao1986generate,goldreich1998secure}, which provides high security at the cost of expensive cryptographic operations, \ac{FL} allows for more efficient implementation and reduced running costs due to its relaxed security requirements. As explicit data exchange is not required, \ac{FL} does not necessitate the addition of noise to the data as in \ac{DP}~\cite{hao2019towards,xu2019ganobfuscator,zhao2019differential,byrd2020differentially,truex2019hybrid}, nor the encryption of data into a homomorphic phase for homomorphic operations as in \ac{HE}~\cite{hardy2017private,aono2017privacy,zhang2020batchcrypt,fang2021privacy}. Gradient aggregation from local models constitutes a core research problem in \ac{FL}. Mcmahan \emph{et al.}~\cite{mcmahan2017communication} presented the \ac{FedAvg} method for training deep neural networks across multiple parties, whereby the global model averages gradients from local models, \emph{i.e.} $\omega' = \sum_{i}^N \frac {1}{N} \omega_i \label{FedAvg}$, with $\omega'$ and $\omega_{i}$ representing the gradients of the global and $i_{th}$ local models, respectively, and $N$ denoting the total number of clients. The method was evaluated on the MNIST benchmark, demonstrating its feasibility for classic image classification tasks using \ac{CNN} as the learning model. Although the experimental results indicate that \ac{FedAvg} is suitable for both \ac{IID} and \ac{Non-IID} data, the statistical challenge of \ac{FL} remains when local models are trained on large-scale \ac{Non-IID} data. Our experimental results support this assertion, as the prediction accuracy and convergence rate significantly decline with large-scale \ac{Non-IID} data when using \ac{FedAvg}.

\ac{FL} is designed for privacy-preserving training, as data is retained and processed locally. Nevertheless, multiple studies, such as~\cite{hitaj2017deep,zhu2019deep,ren2022grnn}, have underscored that \ac{FL} is susceptible to the gradient leakage problem, whereby private training data can be recovered from publicly shared gradients with a considerably high success rate. Hitaj \emph{et al.}~\cite{hitaj2017deep} proposed a training data recovery approach for \ac{FL} systems using \ac{GAN}s. This approach aims to generate training samples similar to a specific class, rather than directly recovering the original training data. Initially, the global \ac{FL} model is trained for several iterations to achieve relatively high accuracy. The authors assume that a malicious participant can obtain a client model, which is then used as a discriminator. Subsequently, an image generator is updated based on the output of the discriminator for a targeted image class. Ultimately, the well-trained generator can produce image samples resembling the training data for the specified image class. Zhu \emph{et al.}~\cite{zhu2019deep} framed the data recovery task as a gradient regression problem, treating pixel values of the input image as random variables optimized using back-propagation while the shared model parameters remain fixed. The objective function measures the Euclidean distance between the shared gradient in \ac{FL} and the gradient generated by the random image input, which is minimized during the training phase. They posited that the optimized input would closely resemble the original training image stored on the local client. Experimental results on public benchmark datasets substantiate this hypothesis, thereby indicating that gradient sharing could potentially result in privacy data leakage. Our prior work, \ac{GRNN}~\cite{ren2022grnn}, further enhances the success rate of leakage attacks using generative models for data recovery, particularly when employing a large batch size in training.

%=====================================================================
\section{Proposed Method}
\label{sec:pm}

\subsection{FedBoosting Framework}

The \ac{FedAvg} method~\cite{mcmahan2017communication} generates a new model by averaging gradients from local clients. Nonetheless, with \ac{Non-IID} data, the weights of local models may converge in divergent directions due to disparities in data distribution. As a result, simple averaging schemes perform inadequately, particularly in the presence of strong bias and extreme outliers~\cite{zhao2018federated,li2019convergence,xu2022acceleration}. Consequently, we propose the use of a boosting scheme, specifically \ac{FedBoosting}, to adaptively combine local models based on their generalization performance on distinct local validation datasets. Simultaneously, to preserve data privacy, information exchange among decentralized clients and the server is restricted. Instead of sharing data between clients, encrypted local models are exchanged through a centralized server and validated independently on each client. Further details can be found in Figure~\ref{fig:FedBoost_illustration}.

\begin{algorithm}[ht!]
 \small
 \caption{\ac{FedBoosting} with \ac{HE} and \ac{DP}: Server}
 % \begin{multicols}{2}
 \begin{algorithmic}[1]
  \STATE build model and initialize weights $\omega_{0}$;
  \FOR{each round $r = 1,2,...,R$}
   \FOR{each client $i \in C_N$}
    \IF{$r==1$}
     \STATE $g_{r}^{*i}, T^i_r \leftarrow Train(r, i, \omega_{r-1})$ via Algorithm~\ref{alg:FedBoost_HE_DP_client}.a; 
     % \COMMENT{Local training.}
    \ELSE
     \STATE $g_{r}^{*i}, T^i_r \leftarrow Train(r, i, G_{r-1}^*)$ via Algorithm~\ref{alg:FedBoost_HE_DP_client}.a;
    \ENDIF
   \ENDFOR
   \FOR{each client $i \in C_N$}
    \STATE generate $\hat{G}_r^{*i}$ via Equ.~(\ref{equ:FedBoost_dp}); 
    % \COMMENT{Generate encrypted whole local gradients.}
    \STATE $V^i_r \leftarrow \sum_{j}^N Evaluate(j, \hat{G}_r^{*i})$ via Algorithm~\ref{alg:FedBoost_HE_DP_client}.b; 
    % \COMMENT{Evaluate local model and obtain validation results.}
   \ENDFOR
   \STATE generate $p_{r}^i$ via Equ.~(\ref{equ:boosting_p}\&\ref{equ:matrix}); 
   % \COMMENT{Compute proportion for each local gradients.}
   \STATE generate $G^*_r$ via Equ.~(\ref{equ:FedBoost_merge}); 
   % \COMMENT{Aggregate local gradients and get the final global gradients.}
   \IF{$r==R$}
    \STATE $\omega_r \leftarrow Decrypt(G^*_r)$ via Algorithm~\ref{alg:FedBoost_HE_DP_client}.c 
    % \COMMENT{Decode global gradients.}
   \ENDIF
  \ENDFOR
 \end{algorithmic}
 % \end{multicols}
 \label{alg:FedBoost_HE_DP_server}
\end{algorithm}

\begin{algorithm}[ht!]
 \small
 \caption{\ac{FedBoosting} with \ac{HE} and \ac{DP}: Client}
 \textbf{a.Train($r, i, \omega_{r-1} || G_{r-1}^*$):}
 \begin{algorithmic}[1]
  \IF{$i==1$}
   \STATE generate key pair and sent to other clients
  \ELSE
   \STATE wait for key pair from $C_1$
  \ENDIF
  \IF{$r==1$}
   \STATE load $TrnD^{i}, ValD^{i}$
  \ELSE
   \STATE decrypt $G_{r-1}^*$ to $G_{r-1}$ by secret keys
   \STATE $\omega_{r-1}=\omega_{r-2}+G_{r-1}$
  \ENDIF
   \FOR{each epoch $e = 1,2,...,E$}
    \FOR{each batch $b = 1,2,...,B$}
     \STATE $\omega_{r} \leftarrow \omega_{r-1} - \eta \cdot \nabla f(TrnD^{i,b}, \omega_{r-1})$
    \ENDFOR
   \ENDFOR
  \STATE $G^i_r=\omega_{r}-\omega_{r-1}$
  \STATE $g^i_r=\lfloor (G^i_r*1e32)/P \rceil$ and generate $g^{*i}_r$ by public keys
  % \STATE encrypt $g^i_r$ to $g^{*i}_r$ by public keys
  \STATE $T^i_r \leftarrow f(TrnD^i | \omega_{r})$
  \STATE return $g^{*i}_r, T^i_r$ to server
 \end{algorithmic}
 \textbf{b.Evaluate($j, \hat{G}_r^{*i}$):}
 \begin{algorithmic}[1]
  \STATE decrypt $\hat{G}_r^{*i}$ to $\hat{G}_r^{i}$ by secret keys
  \STATE $\hat{\omega}_r^i=\omega_{r-1}+\hat{G}_r^{i}$
  \STATE $V^{i,j}_r \leftarrow f(ValD^{j} | \hat{\omega}_{r}^i))$
  \STATE return $V^{i,j}_r$ to server
 \end{algorithmic}
 \textbf{c. Decrypt($G^*_r$):}
 \begin{algorithmic}[1]
  \STATE decrypt $G^*_r$ to $G_r$ by secret keys
  \STATE $\omega_r=\omega_{r-1}+G_r$
  \STATE return $\omega_r$ to server
 \end{algorithmic}
 % \end{multicols}
 \label{alg:FedBoost_HE_DP_client}
\end{algorithm}

In contrast to \ac{FedAvg}, the proposed \ac{FedBoosting} method accounts for the fitness and generalization performance of each client model, adaptively merging the global model using varying weights for all client models. To accomplish this, three different pieces of information are generated from each client: local gradients $G^i_r$, training loss $T^i_r$, and validation loss $V^{i,i}_r$. Here, $G^i_r$ and $T^i_r$ represent the local gradients and training loss from the $i$-th local model in training round $r$, while $V^{i,i}_r$ refers to the validation loss from the $i$-th local model on the $i$-th local validation dataset in training round $r$. The local gradients $G^i_r$ are subsequently distributed to all other clients via a centralized server. Each client can then obtain cross-validated loss $V^{i,j}_r$, where $i\neq j$. Training and validation losses serve as two metrics for evaluating the predictive performance of the models. A model with a relatively large training loss typically indicates poor convergence and weak generalization capabilities. However, it also implies that the model gradient contains ample information for training. Conversely, a model with low training loss does not necessarily ensure good generalization ability (e.g., over-fitting) and may contain less training information. Therefore, we also consider validation loss. These two losses jointly determine the aggregation weight of a local model contributing to the global model, as depicted in Equation~(\ref{equ:boosting_p}). Then, on the server, all validation results of the $i$-th model are summed, denoted as $V^i_r$, representing the $i$-th model's generalization capacity. Taking convergence into account, a \emph{Softmax} layer is employed with input $T_r$. The outputs, combined with $V^i_r$, are utilized for calculating the aggregation weight $p^i_r$. In the current round of aggregation, the new global gradients $G_r$ can be computed by merging all local gradients $G^i_r$ in accordance with their respective weights $p^i_r$:
\begin{align}
G_r &= \sum_{i}^N p^i_r G^i_r; \forall p^i_r\in[0,1] \label{equ:boosting_merge}, \\
p_r &= softmax(softmax(T_r) \cdot \sum_{j}^N V^j_r) \label{equ:boosting_p}, \\
V^{i,j}_r &=
 \begin{pmatrix}
  V^{1,1}_r & V^{1,2}_r & \cdots & V^{1,j}_r \\
  V^{2,1}_r & V^{2,2}_r & \cdots & V^{2,j}_r \\
  \vdots & \vdots & \ddots & \vdots \\
  V^{i,1}_r & V^{i,2}_r & \cdots & V^{i,j}_r
 \end{pmatrix} \label{equ:matrix}
\end{align}
where $T^i_r$ and $p^i_r$ are training loss and mixture coefficient for the $i$-th local model. 
Moreover, the proposed \ac{FedBoosting} approach is resistant to certain malicious attacks, such as data poisoning. For example, if a malicious client introduces poisoned data into the training set and the tainted local model is aggregated with the same weight as other uncontaminated models, our method can mitigate this issue as the validation scores of the compromised model on other clients will be significantly lower, consequently resulting in a substantially reduced aggregation weight.

%------------------------------------------------------------------------
\subsection{HE Aggregation with Quantized Gradient}

\ac{HE} ensures that computations can be performed on encrypted data as $Enc(A) \bullet Enc(B) = Enc(A * B)$, where ``$\bullet$" represents operations on encrypted data, and ``$*$" represents operations on plaintext data. Since \ac{FedBoosting} involves computing the global model based on local gradients, \ac{HE} is utilized in our method to guarantee secure aggregation and gradient exchange among clients and the server. In \ac{FedBoosting}, local models are trained on each client, and subsequently, local gradients are transmitted to the server, where all local gradients are combined to construct global gradients in each round of aggregation. To preserve gradient information, \ac{FedBoosting} employs the \ac{HE} method, \emph{Paillier}~\cite{paillier1999public}. Upon initiation of training, a pair of \ac{HE} keys are distributed among clients, with the public key being utilized for gradient encryption and the secret key for decryption. After each round of local training, local gradients can be calculated by $G^i_r = \omega^i_r - \omega_{r-1}$, where $\omega_{r-1}$ represents the global weight from the previous round, and $\omega^i_r$ is the weight after training in the current round.

Encrypting $G^i_r$ and directly transmitting it to the server is infeasible, as \emph{Paillier} can only process integer values. To circumvent this issue, we propose converting $G^i_r$ into scaled integer form, denoted as $G^{' i}_r$, by multiplying with $1e^{32}$. Since the weighting scheme at the server side would violate the integer-only constraint for homomorphic computation, we ensure aggregation correctness by dividing $G^{' i}_r$ into $P$ segments and then rounding to an integer according to $g^i_r = \lfloor G^{' i}_r / P \rceil$. This results in negligible precision loss, as only the last few bits are discarded. For instance, with $P=10$, value loss occurs only at the $32$-th bit, whereas for $P=100$, the loss occurs at the $31$-th and $32$-th bits. Finally, $g^i_r$ is encrypted using \emph{Paillier}, and the encrypted $g^{i}_r$ is transmitted to the server. Conversely, the client weight $p_r$ is converted to an integer by multiplying $P$ followed by a rounding operation. In \ac{FedBoosting}, the aggregation weight is computed according to Equ.~(\ref{equ:boosting_p} \& \ref{equ:matrix}). The final encrypted global gradients $G^{*}_r$ can be computed by merging all gradients from clients (see Equ.~(\ref{equ:FedBoost_merge})). The final encrypted global gradients $G^{*}r$ are then sent back to each client for decryption and global weight generation by $\omega_r = \omega{r-1} + G_r$, where $G_r$ is decrypted from $G^{*}_r$. The proposed secure aggregation approach using \ac{HE} with quantized gradient is generalizable and can also be employed for \ac{FedAvg}.

\begin{equation}
\label{equ:FedBoost_merge}
G^{*}_r=\sum_{i}^N \lfloor p^i_r \cdot P\rceil \cdot g^{*i}_r; \forall p^i_r\in[0,1]
\end{equation}

%------------------------------------------------------------------------
\begin{figure*}[ht!]
\centering
 \subfigure[FedAvg Model]{
 \centering
 \includegraphics[width=0.25\linewidth]{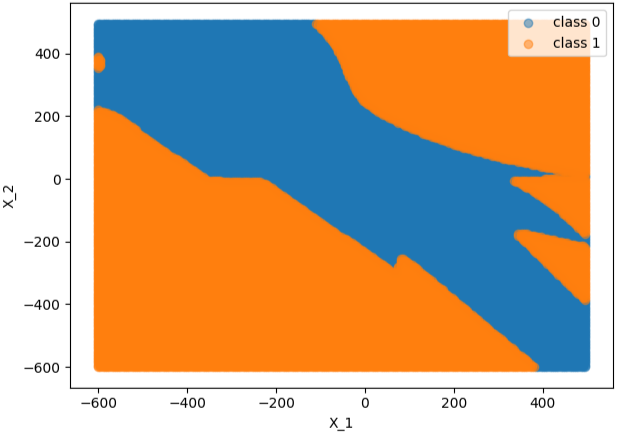}
 }
 \subfigure[FedBoosting Model]{
 \centering
 \includegraphics[width=0.25\linewidth]{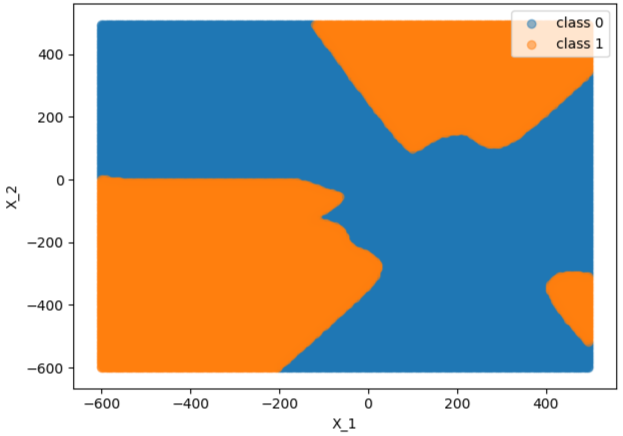}
 }
 \subfigure[Centralized Model]{
 \centering
 \includegraphics[width=0.25\linewidth]{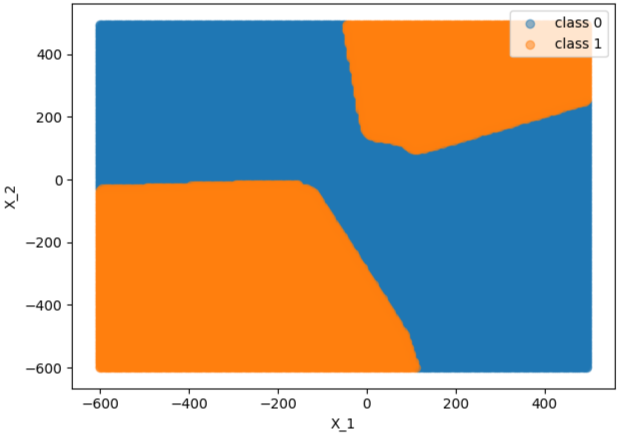}
 }
 \subfigure[Training Dataset One]{
 \centering
 \includegraphics[width=0.25\linewidth]{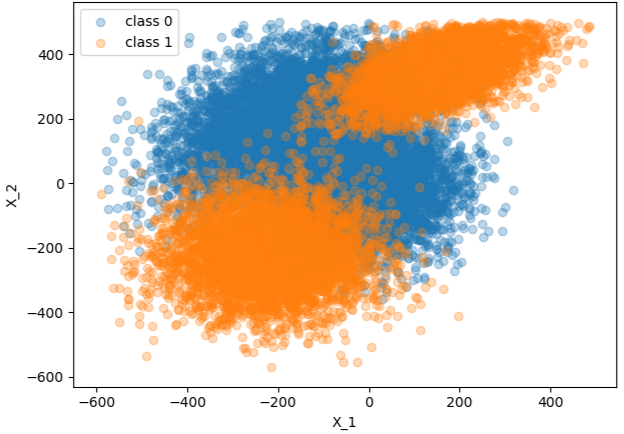}
 }
 \subfigure[Training Dataset Two]{
 \centering
 \includegraphics[width=0.25\linewidth]{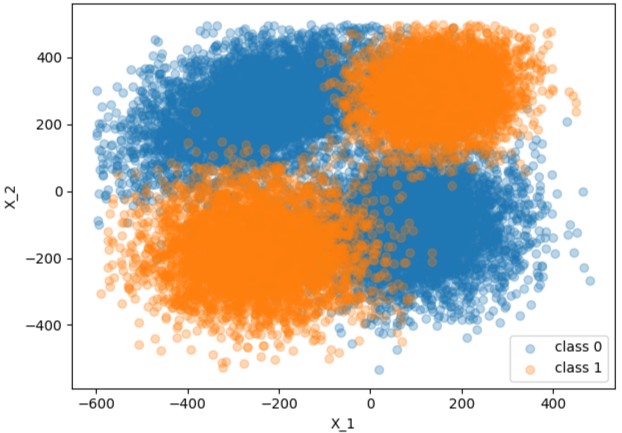}
 }
 \subfigure[Testing Dataset]{
 \centering
 \includegraphics[width=0.25\linewidth]{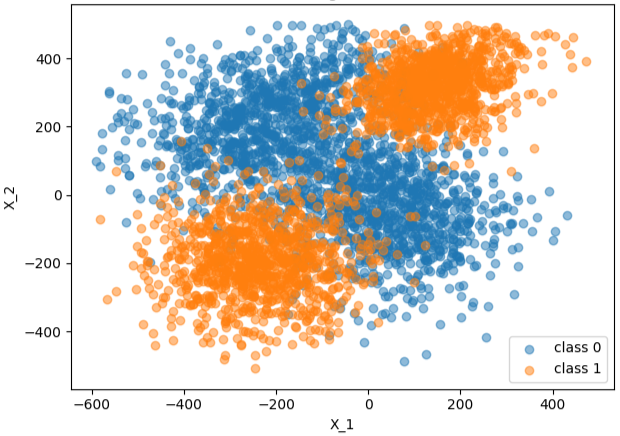}
 }
\caption{Figures (a), (b), and (c) in the first row illustrate the decision boundaries of global models trained employing \ac{FedAvg}, \ac{FedBoosting}, and non-\ac{FL} methods (centralized training scheme), respectively. Figures (d), (e), and (f) in the second row display two training datasets for two clients within an \ac{FL} setting and the testing dataset. The model procured in Figure (c) is trained using all datasets, encompassing both (d) and (e).}
\label{fig:db}
\end{figure*}

\subsection{DP Fusion for Local Model Protection}

The local gradient between the client and server is safeguarded by \ac{HE} as previously mentioned. In the proposed \ac{FedBoosting} mechanism, local models are disseminated to all other clients for cross-validation, and all clients possess an identical key pair. The \ac{FedAvg} demonstrates that the uniformly amalgamated global model is capable of performing comparably to any local model. Thus, in order to protect gradient privacy among clients, we draw inspiration from \ac{DP} and suggest perturbing individual local models through a linear combination with \ac{HE}-encrypted local models, wherein the target model assumes a dominant proportion, yielding the highest weight. Only the perturbed local models are exchanged among clients for cross-validation. Empirically, the reconstructed model exhibits performance akin to that of the local model. Upon receipt of all the encrypted local gradient pieces $g^{*i}_r, \forall i=1,2,...,N$ by the server, the server arbitrarily generates $N$ sets of private fusing weights, within which the corresponding local model consistently assumes the largest proportion. Subsequently, the server computes $N$ reconstructed local models using \ac{HE} in accordance with the $N$ sets of weights (see Equ.~\ref{equ:FedBoost_dp}).
\begin{equation}
\label{equ:FedBoost_dp}
\hat{G}^{*i}_r=\lfloor\hat{p} \cdot P\rceil \cdot g^{*i}_r+\sum_{j:j\ne i}^N\lfloor\frac{(1-\hat{p}) \cdot P}{N-1}\rceil \cdot g^{*j}_r
\end{equation}
where $\hat{G}^{*i}_r$ represents the $i$-th dual-encrypted whole gradient in round $r$. Ultimately, the server disseminates the reconstructed local models to all clients for cross-validation. As \ac{DP} is employed on the server for linear combination and \ac{HE} is utilized on the client, the model is rigorously protected during the exchange process between the server and clients. The performance of the local model may decline due to precision loss resulting from the use of quantized \ac{HE} and linear reconstruction for cross-validation. However, as depicted in Figure~\ref{fig:acc}, our experimental outcomes reveal that there is no substantial loss in testing accuracy. Consequently, the privacy budget $\epsilon$ can be expressed as the logarithm of the performance drop $d$ induced by the linear reconstruction, formulated as $\epsilon = \ln{d}$.

%=====================================================================

\section{Experiments}
\label{sec:er}

\subsection{Decision Boundary Comparison using Synthetic Dataset}
\label{subsec:se}

Initially, we performed evaluations employing two datasets to juxtapose the decision boundary between \ac{FedAvg} and \ac{FedBoosting}. The task encompasses a binary classification problem with 2D features, facilitating a comprehensible visualization of the decision boundary. We posited that the data adheres to a 2D Gaussian distribution, and two datasets were randomly sampled with distinct mean centers and standard deviations to simulate the \ac{Non-IID} scenario. Each individual dataset was utilized for training on a client, and the global model was aggregated using \ac{FedAvg} and the proposed \ac{FedBoosting}, wherein each dataset contained 40,000 samples and was divided into training and testing sets in a 9:1 ratio. Figure~\ref{fig:db} (d), (e), and (f) depict the two training datasets and the combined testing dataset, respectively. A rudimentary neural network was employed, consisting of two fully connected layers followed by a \emph{Sigmoid} activation layer and a \emph{Softmax} layer, respectively. The first fully connected layer contained eight hidden nodes, while the second layer featured two hidden nodes. The \emph{Adam} optimizer with a learning rate of 0.003 was utilized. All models trained via \ac{FedBoosting} surpassed those trained with \ac{FedAvg}, employing a batch size of eight and an epoch of one. Figure~\ref{fig:db} (a) and (b) display visualizations of the decision boundary of global models trained using \ac{FedAvg}, \ac{FedBoosting}, and a centralized training scheme, respectively. It can be inferred that the proposed \ac{FedBoosting} yields a considerably smoother decision boundary compared to the \ac{FedAvg} approach. Moreover, the decision boundary of our method more closely resembles the model trained using a centralized scheme, in which both datasets were employed concurrently. This investigation reveals that our method possesses greater generalizability in principle.

%------------------------------------------------------------------------
\subsection{Evaluation on Text Recognition Task}
We employed \ac{CRNN}\cite{shi2016end} as the local neural network model for the text recognition task. \ac{CRNN} utilizes \emph{VGGNet}\cite{simonyan2014very} as the backbone network for feature extraction, eliminating the fully connected layers and unrolling the feature maps along the horizontal axis. To model the sequential representation, a multi-layer \ac{BiLSTM} network~\cite{graves2008novel} is situated atop the convolutional layers, accepting the unrolled visual features as input and modeling the long-term dependencies within the sequence in both directions. The outputs of \ac{BiLSTM} are channeled into a \emph{Softmax} layer, projecting each element of the unrolled sequence to the probability distribution of potential characters. The character with the highest \emph{Softmax} score is treated as an intermediate prediction. The \ac{CTC}~\cite{graves2006connectionist} decoder is employed to merge intermediate predictions to generate the final output text. For a comprehensive understanding of the \ac{CRNN} model, readers are encouraged to consult the original publication \cite{shi2016end}.

%----------------------------------------------------------
\begin{figure*}[ht]
\centering
\includegraphics[width=0.98\linewidth]{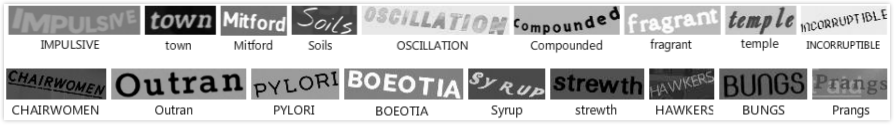}
\includegraphics[width=0.98\linewidth]{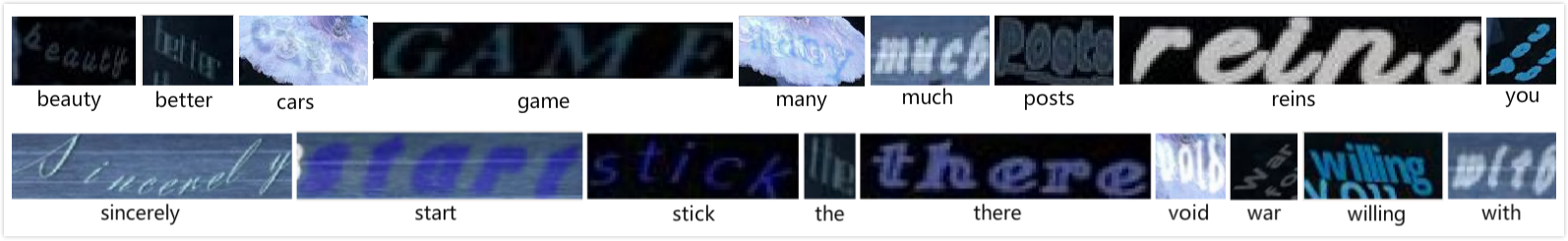}
\caption{The visual example of training images is derived from the \emph{Synth90K} (top two rows) and \emph{SynthText} (bottom two rows) datasets.}
\label{fig:samples}
\end{figure*}

\begin{table*}[ht!]
\setlength{\tabcolsep}{12pt}
\small
\begin{center}
\caption{Recognition accuracies (\%) are presented for four testing datasets. The abbreviations ``90K" and ``ST" represent the \emph{Synth90K} and \emph{SynthText} datasets, respectively. The results in the first row (\ac{CRNN}*) and the second row (\ac{CRNN}) are obtained from the standard CRNN model without employing the \ac{FL} framework. Here, \ac{CRNN}* corresponds to accuracies reported in~\cite{shi2016end}, while \ac{CRNN} denotes the results reproduced using our implementation.}
\label{tab:results}
\begin{tabular}{c|c|c|c|c|c|c|c}
\hline
Method & Dataset & \#Batch & \#Epoch & IIIT5K & SVT & SCUT & IC15 \\
\hline
CRNN* & 90K & - & - & 81.20 & 82.70 & - & - \\
\hline
\multirow{6}{*}{CRNN}
 & ST & 512 & - & 76.07 & 77.60 & 89.38 & 55.92 \\ % epoch 80
 & ST & 800 & - & 73.69 & 78.00 & 86.94 & 58.88 \\ % epoch 90
 & 90K & 512 & - & 80.95 & 86.40 & 87.49 & 67.43 \\ % epoch 70
 & 90K & 800 & - & 80.71 & 83.60 & 87.80 & 62.50 \\ % epoch 70
 & ST \& 90K & 512 & - & 83.81 & 90.40 & 93.08 & 71.71 \\ % epoch 70
 & ST \& 90K & 800 & - & 85.48 & 88.00 & 93.78 & 72.70 \\ % epoch 50
\hline
\multirow{6}{*}{\ac{FedAvg}}
 & ST \& 90K & 256 & 1 & - & - & - & - \\
 & ST \& 90K & 256 & 3 & - & - & - & - \\
 & ST \& 90K & 512 & 1 & 85.48 & 87.60 & 93.31 & 73.36 \\ % round 66
 & ST \& 90K & 512 & 3 & 80.83 & 87.60 & 91.11 & 64.80 \\ % round 60
 & ST \& 90K & 800 & 1 & 86.67 & 89.60 & 94.49 & 72.37 \\ % round 48
 & ST \& 90K & 800 & 3 & 81.82 & 88.00 & 93.47 & 70.07 \\ % round 45
\hline
\multirow{6}{*}{\textbf{Ours}}
  & ST \& 90K & 256 & 1 & 85.83 & 89.2 & 94.26 & 72.37 \\ % round 54
  & ST \& 90K & 256 & 3 & 84.88 & 91.20 & \textbf{94.65} & 70.07 \\ % round 55
  & ST \& 90K & 512 & 1 & 85.12 & 88.00 & 93.39 & 74.34 \\ % round 67
  & ST \& 90K & 512 & 3 & 87.38 & 90.80 & 93.86 & 70.39 \\ % round 47
  & ST \& 90K & 800 & 1 & \textbf{87.62} & 89.20 & 94.18 & \textbf{75.99} \\ % round 45
  & ST \& 90K & 800 & 3 & 85.60 & \textbf{91.60} & \textbf{94.65} & 70.72 \\ % round 44
\hline
\end{tabular}
\end{center}
\end{table*}

%--------------------------------------------------------

\subsubsection{Experimental Setting}

\begin{table*}[ht!]
\small
\begin{center}
\caption{A visual representation of testing outcomes is provided, utilizing \ac{FedAvg} and \ac{FedBoosting} models with a batch size of 512 and an epoch count of 3. Erroneously predicted characters are emphasized in red, and green characters enclosed in brackets signify those omitted from the predictions.}
\label{comparation}
\begin{tabular}{m{0.26\linewidth} m{0.26\linewidth} m{0.26\linewidth}}
\hline
{\bfseries Samples} & {\bfseries \ac{FedAvg}} & {\bfseries \ac{FedBoosting}}\\
\hline
\rule{0pt}{24pt}\includegraphics[width=0.8\linewidth]{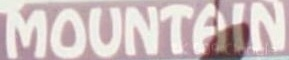} & MOUNT{\color{red} R}IN & MOUNTAIN\\
\hline
\rule{0pt}{28pt}\includegraphics[width=0.8\linewidth]{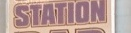} & STATIONN{\color{red} N} & STATION\\
\hline
\rule{0pt}{20pt}\includegraphics[width=0.8\linewidth]{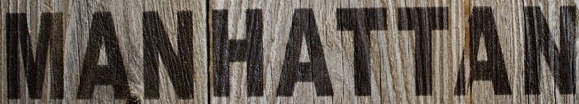} & MANHAT{\color{red} I}TAN & MANHATTAN\\
\hline
\rule{0pt}{20pt}\includegraphics[width=0.8\linewidth]{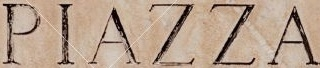} & P{\color{red} L}AZZA & PIAZZA\\
\hline
\rule{0pt}{20pt}\includegraphics[width=0.8\linewidth]{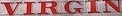} & VIRG({\color{green} I})N & VIRGIN\\
\hline
\rule{0pt}{20pt}\includegraphics[width=0.8\linewidth]{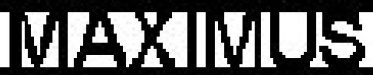} & MAXI{\color{red} V}US & MAXIMUS\\
\hline
\rule{0pt}{20pt}\includegraphics[width=0.8\linewidth]{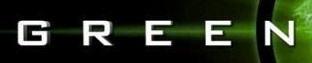} & {\color{red} U}GREEN{\color{red} N} & GREEN\\
\hline
\rule{0pt}{22pt}\includegraphics[width=0.8\linewidth]{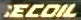} & {\color{red} J}ECOIL & ECOIL\\
\hline
\rule{0pt}{20pt}\includegraphics[width=0.8\linewidth]{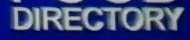} & DIR({\color{green} E})C{\color{red} I}O{\color{red} N} & DIRECTORY\\
\hline
\rule{0pt}{24pt}\includegraphics[width=0.8\linewidth]{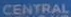} & CEN{\color{red} I}RAL & CENTRAL\\
\hline
\rule{0pt}{22pt}\includegraphics[width=0.8\linewidth]{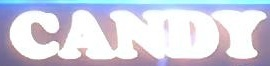} & CAN{\color{red} S}Y & CANDY\\
\hline
\rule{0pt}{20pt}\includegraphics[width=0.8\linewidth]{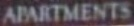} & A{\color{red} B}ARTMENTS & APARTMENTS\\
\hline
\rule{0pt}{28pt}\includegraphics[width=0.8\linewidth]{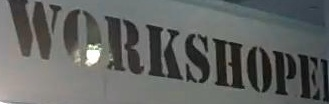} & WORKSHOPH({\color{green} E}) & WORKSHOPE\\
\hline
\rule{0pt}{22pt}\includegraphics[width=0.8\linewidth]{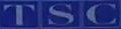} & T{\color{red} I}SC & TSC\\
\hline
\rule{0pt}{28pt}\includegraphics[width=0.8\linewidth]{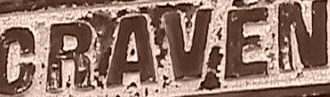} & CR{\color{red} I}AVEN & CRAVEN\\
\hline
\rule{0pt}{24pt}\includegraphics[width=0.8\linewidth]{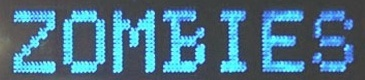} & {\color{red} 20}MBIES & ZOMBIES\\
\hline
\rule{0pt}{26pt}\includegraphics[width=0.8\linewidth]{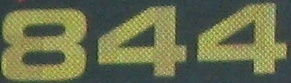} & {\color{red} 3}44 & 844\\
\hline
\rule{0pt}{24pt}\includegraphics[width=0.8\linewidth]{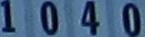} & 1{\color{red} 7}040 & 1040\\
\hline
\rule{0pt}{28pt}\includegraphics[width=0.8\linewidth]{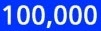} & 100{\color{red} 1}000 & 100000\\
\hline
\rule{0pt}{20pt}\includegraphics[width=0.8\linewidth]{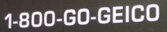} & 1800GOGEICO & 1800G{\color{red} 0}GE{\color{red} 1}CO\\
\hline
% \rule{0pt}{30pt}\includegraphics[width=0.8\linewidth]{comparation/samples_YMCA_.png} & YMCA & {\color{red} I}MCA\\
% \hline
\end{tabular}
\end{center}
\end{table*}

The proposed model is trained on two extensive synthetic datasets, \emph{Synthetic Text 90k} and \emph{Synthetic Text}, without fine-tuning on other datasets. The model is assessed on four additional standard datasets to evaluate its general recognition performance. For all experiments, word-level accuracy serves as the metric for prediction performance.

\textbf{Synthetic Text 90k}~\cite{jaderberg2014synthetic} (Synth90K) constitutes one of two training datasets employed in this paper. This dataset comprises approximately 7.2 million images and their corresponding ground truth words. For \ac{FedBoosting}, we divide the images into two subsets: the first containing 6.5 million images for training, and the second consisting of 0.7 million images for validation.

\textbf{Synthetic Text}~\cite{jaderberg2014synthetic} (SynthText) serves as the second training dataset utilized. This dataset contains approximately 85K natural images featuring numerous synthetic texts. We crop all texts using labeled text bounding boxes to create a new dataset of 5.3 million text images. For \ac{FedBoosting}, we partition this dataset into a training dataset of 4.8 million images and a validation dataset of 0.5 million images.

\textbf{IIIT 5K-Words}~\cite{mishra2012scene} (IIIT5K) is sourced from the internet and encompasses 3000 cropped word images in its testing set, with each image containing a ground truth word.

\textbf{Street View Text}~\cite{wang2011end} (SVT) is derived from \emph{Google Street View} and consists of 647 word images in its testing set. Many images in this dataset are significantly degraded by noise and blur or exhibit extremely low resolutions.

\textbf{SCUT-FORU}~\cite{zhang2016character} (SCUT) includes 813 training images and 349 testing images, characterized by significant variations in background and illumination.

\textbf{ICDAR 2015}~\cite{karatzas2015icdar} (IC15) comprises 2077 cropped images that feature relatively low resolutions and multi-oriented texts. For a fair comparison, we exclude images containing non-alphanumeric characters, resulting in a final count of 1811 images.

\begin{figure*}[ht]
\centering
 \subfigure[IIIT5K]{
 \centering
 \includegraphics[width=0.25\linewidth]{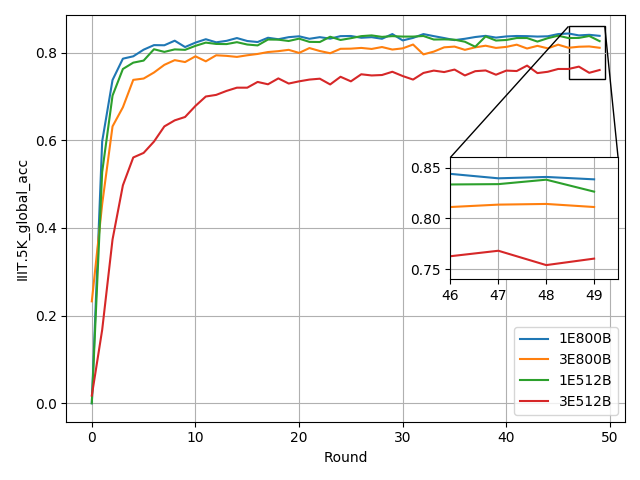}
 }
 \subfigure[SCUT]{
 \centering
 \includegraphics[width=0.25\linewidth]{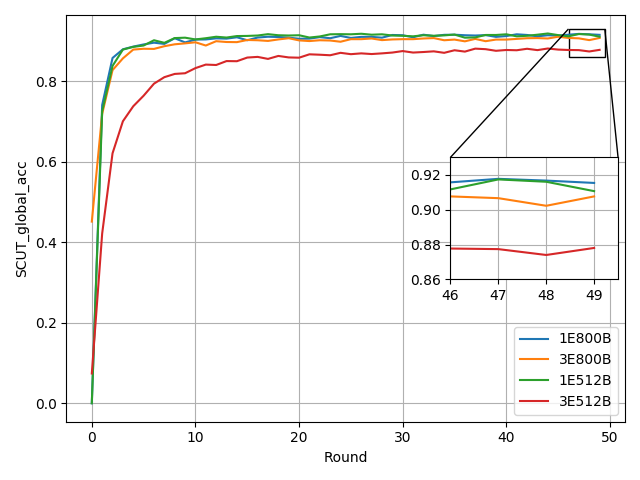}
 }
 \subfigure[ICDAR2015]{
 \centering
 \includegraphics[width=0.25\linewidth]{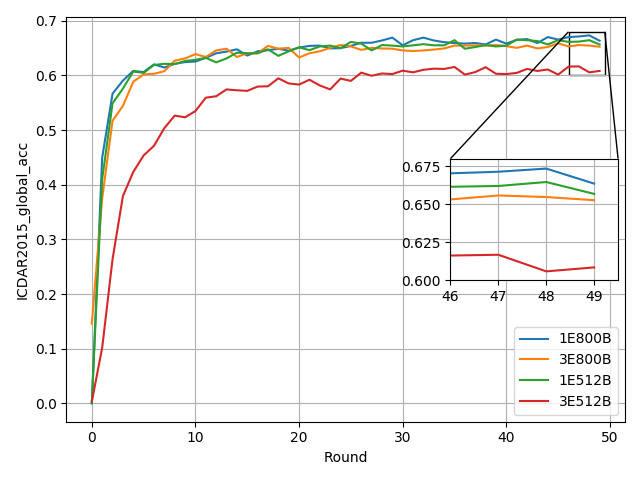}
 }
 \subfigure[IIIT5K]{
 \centering
 \includegraphics[width=0.25\linewidth]{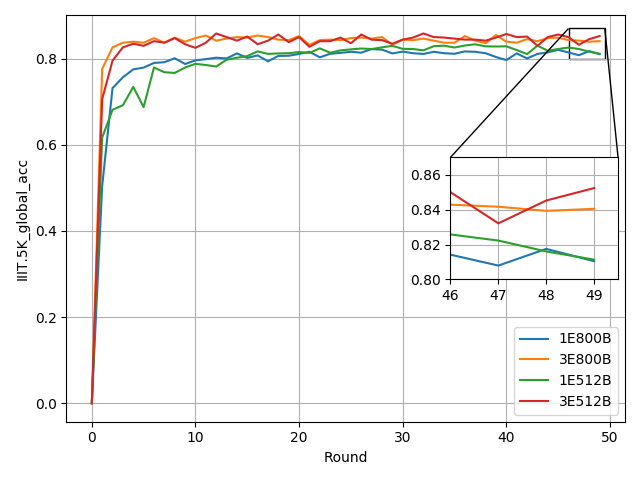}
 }
 \subfigure[SCUT]{
 \centering
 \includegraphics[width=0.25\linewidth]{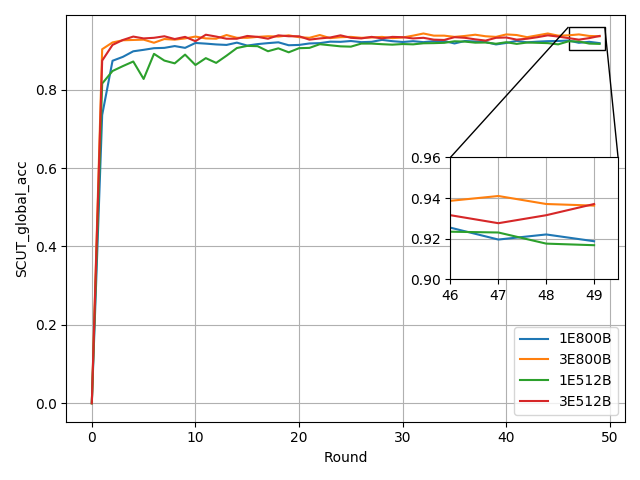}
 }
 \subfigure[ICDAR2015]{
 \centering
 \includegraphics[width=0.25\linewidth]{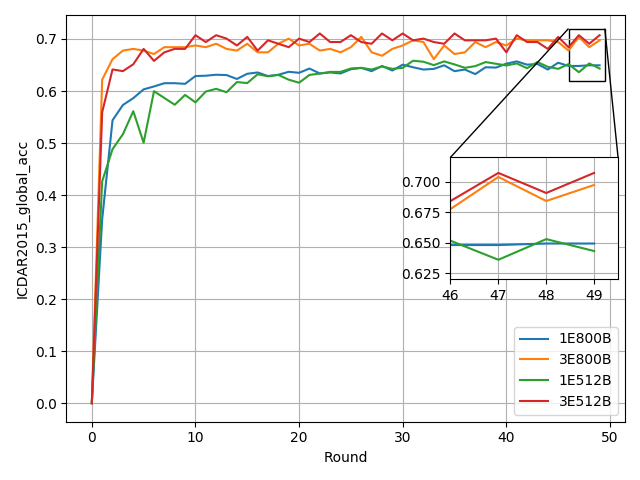}
 }
\caption{Testing accuracies of \ac{FedAvg} (first row) and \ac{FedBoosting} (second row) models across rounds for the \emph{IIIT5K}, \emph{SCUT}, and \emph{IC15} datasets are presented. The notation ''1E800B" denotes that the model is trained on a client with a batch size of 800 and one epoch. All samples in these testing subsets are resized to 100$\times$32, differing from the processing in Table~\ref{tab:results}.}
\label{fig:accuracy}
\end{figure*}

\begin{figure*}[ht]
\begin{center}
\subfigure[256 batch size and 1 epoch]{\includegraphics[width=0.43\linewidth]{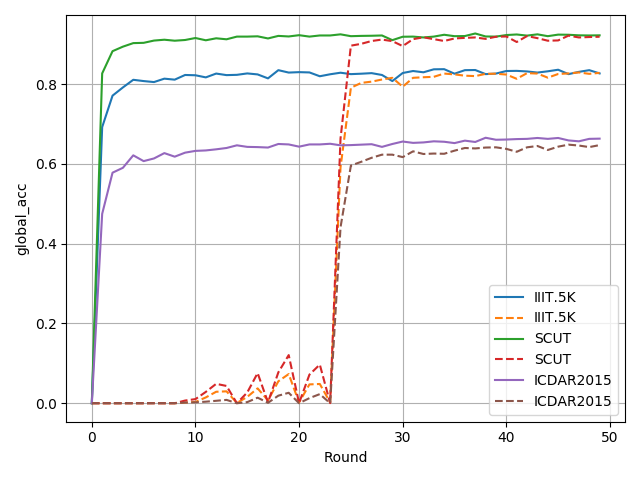}}
\subfigure[800 batch size and 1 epoch]{\includegraphics[width=0.43\linewidth]{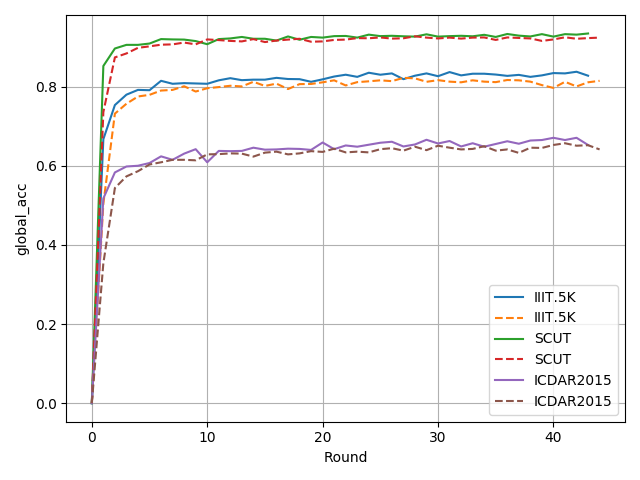}}
\caption{Performance comparison on how the training results affect the model performance. Solid lines refer to global models aggregated with training and validation results, while dash lines refer to global models aggregated only with validation results.}
\label{fig:pc}
\end{center}
\end{figure*}

Figure~\ref{fig:samples} presents visual instances from the \emph{Synth90K} and \emph{SynthText} datasets, illustrating considerable variations in backgrounds and texts between the two sets. Consequently, it can be deduced that these datasets exhibit \ac{Non-IID} characteristics. To accommodate mini-batch processing and expedite the training procedure, all training and validation images are resized to dimensions of 100$\times$32. In Tables~\ref{tab:results} and \ref{tab:he_results}, testing images are proportionally resized to maintain a height of 32 pixels. Figures~\ref{fig:accuracy}, \ref{fig:acc}, and \ref{fig:pc} display testing images processed identically to the aforementioned training and validation images. Due to the constraints imposed by \ac{CTC}, testing images with label lengths less than 3 or greater than 25 characters are excluded. The \emph{Synth90K} and \emph{SynthText} datasets are deployed on two distinct clients, with \emph{AdaDelta} employed for back-propagation optimization on local training nodes, and an initial learning rate of 0.05. For \ac{HE}, a key size of 128 bits is utilized, and the entire gradient is divided into 100 segments with $\hat{p}=0.9$. Our method is implemented using \emph{Keras} and \emph{Tensorflow} in a distributed multi-\ac{GPU}s setup, with the source code publicly accessible to ensure reproducibility.

%----------------------------------------------------------
\subsubsection{Result and Discussion}
\label{subsec:results}

Table~\ref{tab:results} presents a comparative analysis of test dataset outcomes using various training hyperparameters, including batch size and number of epochs. The first row (\ac{CRNN}*) and the second row (\ac{CRNN}) represent results produced by the original \ac{CRNN} model without employing the \ac{FL} framework, where \ac{CRNN}* corresponds to the accuracies reported by its authors in~\cite{shi2016end}, and \ac{CRNN} corresponds to the results we reproduced. In comparison to the original \ac{CRNN} model, \ac{FedAvg} exhibits a substantial improvement. For instance, the \ac{FedAvg} model with a batch size of 800 and an epoch of 1 achieves an accuracy of 86.67\% on the \emph{IIIT5k} dataset, signifying a 1.19\% enhancement compared to the \ac{CRNN} result of 85.48\% using the same configuration. The \ac{FedAvg} model with a batch size of 512 and an epoch of 1 exhibits a 1.65\% improvement on the \emph{IC15} dataset. More significantly, the proposed \ac{FedBoosting} method attains the highest accuracy across all four test datasets, with 87.62\%, 91.60\%, 94.65\%, and 75.99\% reported on \emph{IIIT5k}, \emph{SVT}, \emph{SCUT}, and \emph{IC15}, respectively. Additionally, our approach outperforms both \ac{FedAvg} and non-\ac{FL} methods by considerable margins. Table~\ref{comparation} provides more qualitative results.

It can be observed that \ac{FedAvg} models with larger batch sizes and fewer epochs exhibit superior performance. In other words, the models perform better when model integration occurs more frequently, albeit at the expense of increased communication costs. In Table~\ref{tab:results}, the model with a batch size of 256 and one epoch does not produce a result due to model divergence after several integration rounds. The potential reason could be the extreme differences in parameters learned on each local machine. Figure~\ref{fig:accuracy} compares the convergence curves of \ac{FedAvg} and the proposed \ac{FedBoosting}. The convergence curves of \ac{FedAvg} models with smaller batch sizes or more epochs are consistently lower than those of models with larger batch sizes or fewer epochs. For instance, employing the \ac{FedAvg} strategy, the model with a batch size of 800 and one epoch (iterating 6,689 and 9,030 times per epoch on the \emph{SynthText} and \emph{Synth90K} datasets, respectively) performs significantly better than the model with a batch size of 512 and three epochs (iterating 10,452 and 14,110 times per epoch on the \emph{SynthText} and \emph{Synth90K} datasets, respectively) on the \emph{IIIT5K} test dataset. Conversely, the accuracy curves of \ac{FedBoosting} models (as seen in Figure~\ref{fig:accuracy}, second row) do not exhibit this issue. Consequently, we deduce that the proposed boosting strategy effectively mitigates the model collapse issue of \ac{FedAvg} to a significant extent.

\begin{figure*}[ht!]
\centering
 \subfigure[IIIT5K]{
 \centering
 \includegraphics[width=0.25\linewidth]{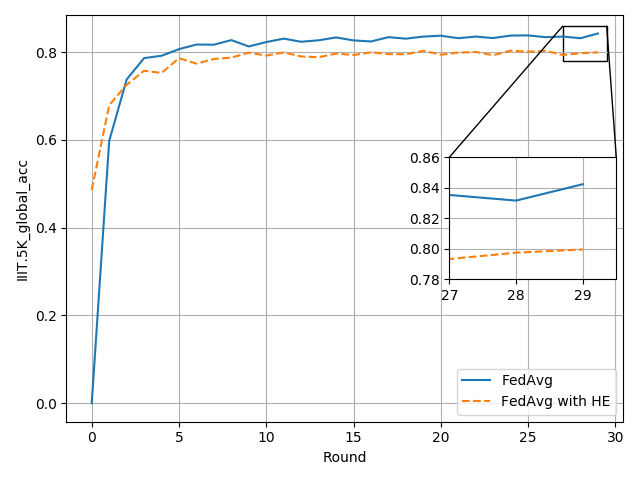}
 }
 \subfigure[SVT]{
 \centering
 \includegraphics[width=0.25\linewidth]{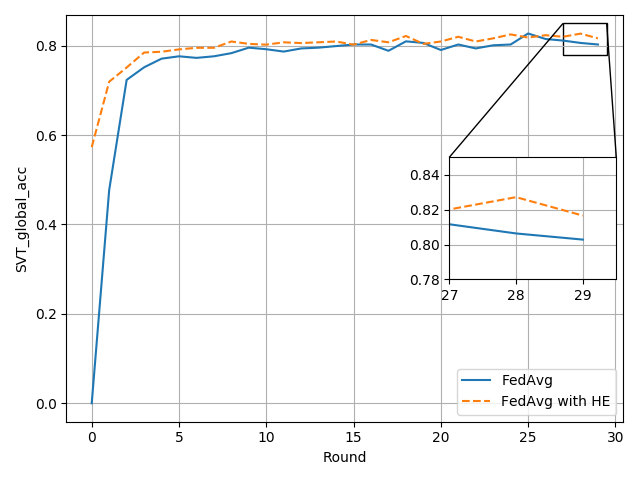}
 }
 \subfigure[ICDAR2015]{
 \centering
 \includegraphics[width=0.25\linewidth]{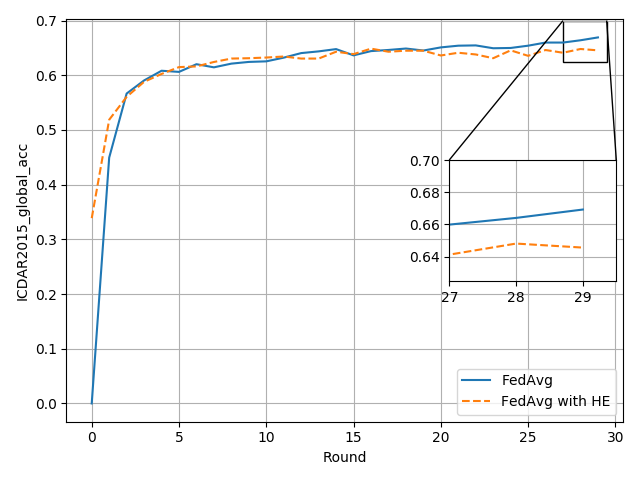}
 }
 \subfigure[IIIT5K]{
 \centering
 \includegraphics[width=0.25\linewidth]{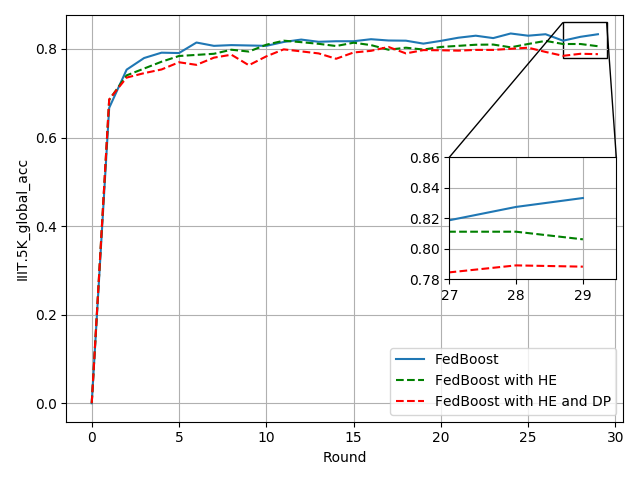}
 }
 \subfigure[SVT]{
 \centering
 \includegraphics[width=0.25\linewidth]{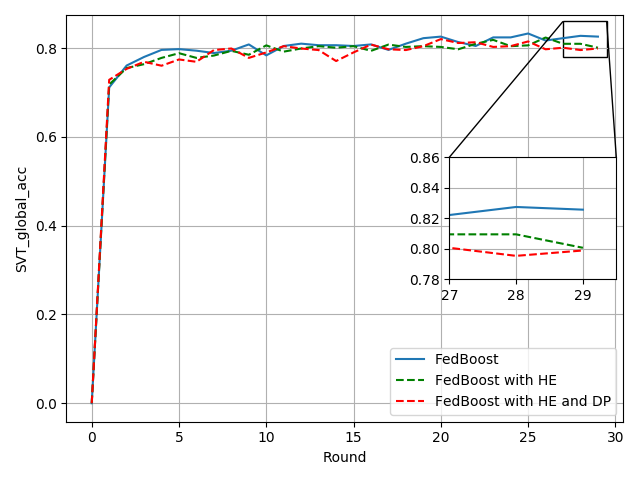}
 }
 \subfigure[ICDAR2015]{
 \centering
 \includegraphics[width=0.25\linewidth]{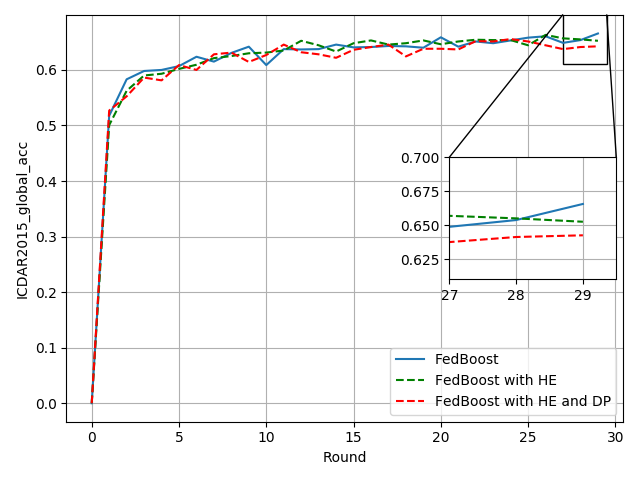}
 }
\caption{Testing accuracy of \ac{FedAvg} (first row) and \ac{FedBoosting} (second row) models with and without using encryption protocol over training rounds on \emph{IIIT5K}, \emph{SVT} and \emph{IC15} datasets.}
\label{fig:acc}
\end{figure*}

\begin{table}[ht!]
\small
\setlength{\tabcolsep}{8pt}
\begin{center}
\caption{Recognition accuracies (\%) on three testing datasets. All experiments are using batch size of 800 and epoch of 1.}
\label{tab:he_results}
\begin{tabular}{c|c|c|c|c}
\hline
Method & Encryption & IIIT5K & SVT & IC15 \\
\hline
\multirow{2}{*}{\ac{FedAvg}}
 & N/A & 86.67 & \textbf{89.60} & 72.37 \\ % round 48
 & \ac{HE} & 86.40 & 88.00 & 73.00 \\ % round 34 
\hline
\multirow{3}{*}{\ac{FedBoosting}}
 & N/A & \textbf{87.62} & 89.20 & \textbf{75.99} \\ % round 45
 & \ac{HE} & 85.12 & 88.40 & 72.70 \\ % round 23
 & \ac{HE}+\ac{DP} & 85.00 & 88.80 & 72.37 \\ % round 25
\hline
\end{tabular}
\end{center}
\end{table}

\begin{table}[ht!]
\small
\setlength{\tabcolsep}{8pt}
\begin{center}
\caption{Validation accuracies from local models without and with linear reconstruction.}
\label{tab:dp_impact}
\begin{tabular}{c|c|c|c|c}
\hline
 Client & \#Batch/\#Epoch & w/o \ac{DP} & w/ \ac{DP} & Gap \\
\hline
 1 & \multirow{2}{*}{256/1} & 94.34\% & 92.23\% & -2.11\% \\
 2 &                        & 90.32\% & 86.83\% & -3.49\% \\
\hline
 1 & \multirow{2}{*}{256/3} & 95.81\% & 92.82\% & -2.99\% \\
 2 &                        & 90.26\% & 86.92\% & -3.34\% \\
\hline
 1 & \multirow{2}{*}{512/1} & 94.09\% & 92.31\% & -1.78\% \\
 2 &                        & 88.34\% & 85.29\% & -3.05\% \\
\hline
 1 & \multirow{2}{*}{512/3} & 96.46\% & 92.89\% & -3.57\% \\
 2 &                        & 91.84\% & 86.71\% & -5.13\% \\
\hline
 1 & \multirow{2}{*}{800/1} & 94.71\% & 92.52\% & -2.19\% \\
 2 &                        & 89.44\% & 87.19\% & -2.25\% \\
\hline
 1 & \multirow{2}{*}{800/3} & 97.08\% & 92.74\% & -4.34\% \\
 2 &                        & 93.45\% & 87.91\% & -5.54\% \\
\hline
\end{tabular}
\end{center}
\end{table}

Table~\ref{tab:he_results} presents a comparative analysis of three testing datasets (\emph{IIIT5K}, \emph{SVT}, and \emph{IC15}) using various \ac{FL} gradient merging techniques and encryption modes, given hyperparameters of 800 batch size and 1 epoch. The results for \ac{FedAvg} indicate that, although employing \ac{HE} results in a minor precision loss due to the partitioning of the entire gradient into multiple pieces, accuracy is virtually unaffected on the testing datasets. In fact, there is an increase in accuracy on the \emph{IC15} dataset from 72.37\% to 73.00\%. On the remaining two testing datasets, the accuracy losses are 0.27\% and 1.6\%, respectively. Similarly, for \ac{FedBoosting} models, testing outcomes exhibit a slight accuracy reduction, which can be tolerated when using \ac{HE} exclusively. When incorporating \ac{DP} with \ac{HE} in \ac{FedBoosting}, the accuracy of the global model remains largely unaffected. This is because \ac{DP} encryption is only employed to encrypt local gradients between clients for evaluation and to obtain results on all clients' validation datasets. Table~\ref{tab:dp_impact} illustrates the impact of \ac{DP} on local model performance. Validation experiments using \ac{DP} are conducted on individual clients, necessitating the incorporation of \ac{HE}. We hypothesize that \ac{HE} has a minimal effect on validation accuracy since it only modifies the final two digits out of 32 decimal places. The disturbance to $p_r$ is relatively insignificant, as referred to in Equation~\ref{equ:boosting_p}, leading to the conclusion that \ac{DP} has limited influence on the generation of global gradients. Nevertheless, testing results exhibit a decline between standard \ac{FedBoosting} and encrypted \ac{FedBoosting} models, for example, accuracy is reduced from 87.62\% to approximately 85.00\% on \emph{IIIT5K} and around 3.00\% on \emph{IC15}. This is a typical fluctuation for training \ac{DL} models. Although testing accuracies for all three datasets exhibit varying degrees of reduction, the accuracy growth trends are represented by the differential curves under different encryption modes in Figure~\ref{fig:acc}. It can be observed that the differentials for most testing datasets are relatively small.

%----------------------------------------------------------

\subsubsection{Performance Comparison}
\label{subsec:as}

We further posit that the divergence issue can be attributed to the quality of the datasets, as illustrated in Figure~\ref{fig:samples}. In other words, each local model trained on distinct private datasets undoubtedly possesses varying generalization capabilities. In our experiments, the crude aggregation of the global model by averaging the weights of local models may lead to a decline in generalization ability, particularly when the number of local updating iterations is large (i.e., small batch size or large epoch number). Consequently, the proposed \ac{FedBoosting} aims to assign a more equitable weight instead of a mean value by balancing the training and validation performance of a local model. Guided by this rationale, \ac{FedBoosting} initially takes into account each model's validation results on every client's validation dataset and subsequently incorporates training results to compute the weights of local models. The consideration of training results stems from the fact that a local model trained on a high-quality dataset typically exhibits a satisfactory fitness, which may, however, perform poorly on low-quality validation datasets. It would be unjust to deem this model as having inadequate generalization ability based solely on its validation performance across datasets of varying quality. Conversely, a model trained on a poor-quality dataset might excel on a high-quality validation dataset; however, we do not desire this type of local model to dominate the global model. As a result, to balance the performance of a local model, we initially aggregate the validation results as a reference representing the local model's generalization ability. Moreover, training results are factored in to adjust the reference and derive the final weights for each local model. Our experiments reveal that the weights amount to approximately 55.00\% for the local model trained on the \emph{Synth90K} dataset and 45.00\% on the \emph{SynthText} dataset, which is reasonable considering the accuracy results in Table~\ref{tab:results} that \ac{CRNN} models trained on the \emph{Synth90K} dataset consistently outperform those trained on the \emph{SynthText} dataset. If we disregard training results, the weight for the local model trained on the \emph{Synth90K} dataset would be smaller than that on the \emph{SynthText} dataset.

To prove the above idea in \ac{FedBoosting}, a performance comparison is given here. It is commonly accepted that generalization ability is a good metrics of judging a model's performance, whereas only considering generalization ability is not feasible for our proposed method \ac{FedBoosting}. Otherwise, it is impossible as well to deploy our method only considering training results and get rid of validation results, which may lead to an extremely unfair situation that local model trained on \emph{Synth90K} may take a weight up to about 80\% for the global model. So the following content is mainly talking about how training results work in \ac{FedBoosting}. We trained a global model with 256 batch size and 1 epoch under the strategy of \ac{FedBoosting} without considering training results. As described above, the reason of thinking over training results is to rectify the weight for local model. From Figure~\ref{fig:pc} (a), we can see that the global model without taking training results gains a delay convergence at round 24. While in other experiments, models all converge quickly and properly under the supervision of training results. In the meantime, the performance of global model with training results is always better than that without training results. As a supplement, we visualize the global testing accuracy of two models with 800 batch size and 1 epoch in Figure~\ref{fig:pc} (b), one uses training results and the other one does not. Two models converge normally in this case, but the model performance of using training results outperforms all the time. From all above, we consider that using training results to supervise the local weight is essential in our scenario. To clarify, all testing images during training are resized to 100$\times$32, which is different to individual testing experiments where testing images are resized to W$\times$32, where W is the proportionally scaled with heights, but at least 100 pixels. That is why accuracies in Figure~\ref{fig:pc} are lower than those in Table~\ref{tab:results}. Please refer to our codes for more details.

In order to substantiate the aforementioned concept within the \ac{FedBoosting} framework, a comparative analysis of performance is presented herein. It is widely acknowledged that a model's generalization ability serves as a reliable metric for evaluating performance; however, relying solely on generalization ability is not tenable for our proposed \ac{FedBoosting} methodology. Likewise, implementing our approach by solely taking into account training outcomes while disregarding validation results may lead to an egregiously unjust scenario in which a local model trained on \emph{Synth90K} might contribute up to approximately 80.00\% of the weight for the global model. Consequently, the subsequent discussion primarily focuses on the role of training outcomes in \ac{FedBoosting}. We trained a global model employing a 256 batch size and a single epoch under the \ac{FedBoosting} strategy, without considering training results. The impetus for incorporating training outcomes is to adjust the local model's weight. As depicted in Figure~\ref{fig:pc} (a), a global model that excludes training results exhibits delayed convergence at round 24. Conversely, in other experiments, models consistently exhibit rapid and appropriate convergence under the guidance of training results. Simultaneously, the performance of the global model that incorporates training results consistently surpasses that of the model which does not. Additionally, we provide a visualization of the global testing accuracy for two models with an 800 batch size and a single epoch in Figure~\ref{fig:pc} (b); one model employs training results while the other does not. Both models demonstrate normal convergence in this instance, but the performance of the model utilizing training results consistently excels. Based on the aforementioned evidence, we posit that supervising local weight with training results is indispensable in our context. For clarity, all testing images during training are resized to 100$\times$32, which differs from individual testing experiments where testing images are resized to W$\times$32, where W is proportionally scaled with heights but maintains a minimum of 100 pixels. This discrepancy accounts for the lower accuracies observed in Figure~\ref{fig:pc} compared to those in Table~\ref{tab:results}. Further details can be found in our code documentation.

%=====================================================================
\section{Conclusion}
\label{sec:cc}

In the present study, we introduced \ac{FedBoosting}, a boosting scheme for the \ac{FL} framework, to address the limitations of the \ac{FedAvg} algorithm on \ac{Non-IID} datasets. To safeguard against gradient leakage attacks, a gradient-sharing protocol incorporating \ac{HE} and \ac{DP} was devised. A thorough comparative analysis was conducted using synthetic datasets and public text recognition benchmarks, demonstrating superior performance compared to traditional \ac{FedAvg} schemes on \ac{Non-IID} datasets. Our implementation is publicly accessible to ensure reproducibility and is compatible with distributed multi-\ac{GPU}s setups. Potential areas for future exploration include the theoretical examination of model convergence within multi-party computing, privacy leakage risks associated with gradients, and the development of more efficient gradient quantization methods.

%=====================================================================
\section*{Acknowledgment}
This work was supported by the Engineering and Physical Sciences Research Council (EPSRC) UK, Data Release - Trust, Identity, Privacy and Security [Grant Number: EP/N028139/1].
%=====================================================================

\bibliographystyle{ieeetr}
\bibliography{bib/ref}

\end{document}